\documentclass{article}

\usepackage[nonatbib,dblblindworkshop,final]{neurips_2025}
\usepackage[square,numbers,compress]{natbib}

\workshoptitle{MATH-AI}

\usepackage[T1]{fontenc}    
\usepackage[colorlinks=true, pdfborder={0 0 0}, linkcolor=blue, citecolor=blue, urlcolor=blue]{hyperref}       
\usepackage{url}            
\usepackage{booktabs}       
\usepackage{amsfonts}       
\usepackage{nicefrac}       
\usepackage{microtype}      
\usepackage[table]{xcolor}
\usepackage{amsmath} 
\usepackage{caption}

\usepackage[most,skins,theorems]{tcolorbox}
\usepackage{wrapfig}

\usepackage{times}
\usepackage{latexsym}
\usepackage{amsmath, amssymb, amsfonts} 
\usepackage{algorithm}
\usepackage[noend]{algpseudocode}

\usepackage{graphicx}
\usepackage{fix-cm}
\usepackage{booktabs} 
\usepackage{tabularx} 
\usepackage{ragged2e} 
\usepackage{siunitx}    

\usepackage{caption} 
\usepackage{pgfplots} 
\definecolor{custom_blue}{HTML}{97d4ff} 
\definecolor{custom_pink}{HTML}{ffa9db} 
\definecolor{custom_green}{HTML}{a9ffcd} 
\definecolor{custom_orange}{HTML}{ffcda9} 

\title{SAND-Math: Using LLMs to Generate Novel, Difficult and Useful Mathematics Questions and Answers}

\author{%
  Chaitanya Manem $^1$\thanks{Corresponding Author}  
  \quad \textbf{Pratik Prabhanjan Brahma} $^1$
  \quad Prakamya Mishra $^1$ \\
  \textbf{Zicheng Liu} $^1$
  \quad \textbf{Emad Barsoum} $^1$ \\
  $^1$Advanced Micro Devices, Inc. (AMD)
}

\begin{document}

\maketitle

\begin{center}
\vspace{-1cm}
\begin{tabular}{rl}
\includegraphics[height=1.05em]{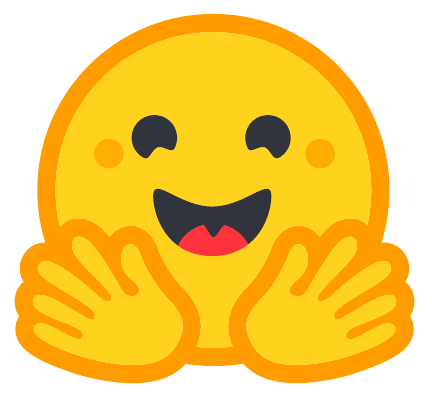} & \url{https://huggingface.co/datasets/amd/SAND-MATH}\\
\end{tabular}
\end{center}
\begin{abstract}
  The demand for Large Language Models (LLMs) at multiple scales, capable of sophisticated and sound mathematical reasoning, continues to grow. However, the development of performant mathematical LLMs is often bottlenecked by the scarcity of useful training data containing problems with significant complexity. We introduce \textbf{SAND-Math} (\textbf{S}ynthetic \textbf{A}ugmented \textbf{N}ovel and \textbf{D}ifficult Mathematics problems and solutions), a pipeline that addresses this by first synthesizing high-quality problems from scratch and then systematically elevating their complexity via a our newly proposed \textbf{Difficulty Hiking} step. We demonstrate the effectiveness of our approach through two key findings: \textbf{(1)} Augmenting a strong post-training baseline with a small 500-sample SAND-Math dataset significantly boosts performance, outperforming the next-best synthetic dataset by $\uparrow$ 17.85 absolute points on AIME25 benchmark. \textbf{(2)} In a dedicated ablation study, we show the effectiveness of our Difficulty Hiking process in increasing average problem difficulty from 5.02 to 5.98. This step consequently lifts AIME25 results from 46.38\% to 49.23\%. The full generation pipeline, final dataset, and a fine-tuned model form a practical and scalable toolkit for building capable and efficient mathematical reasoning LLMs.
\end{abstract}

\begin{figure*}[htbp]
\centering
  \includegraphics[width=\textwidth]{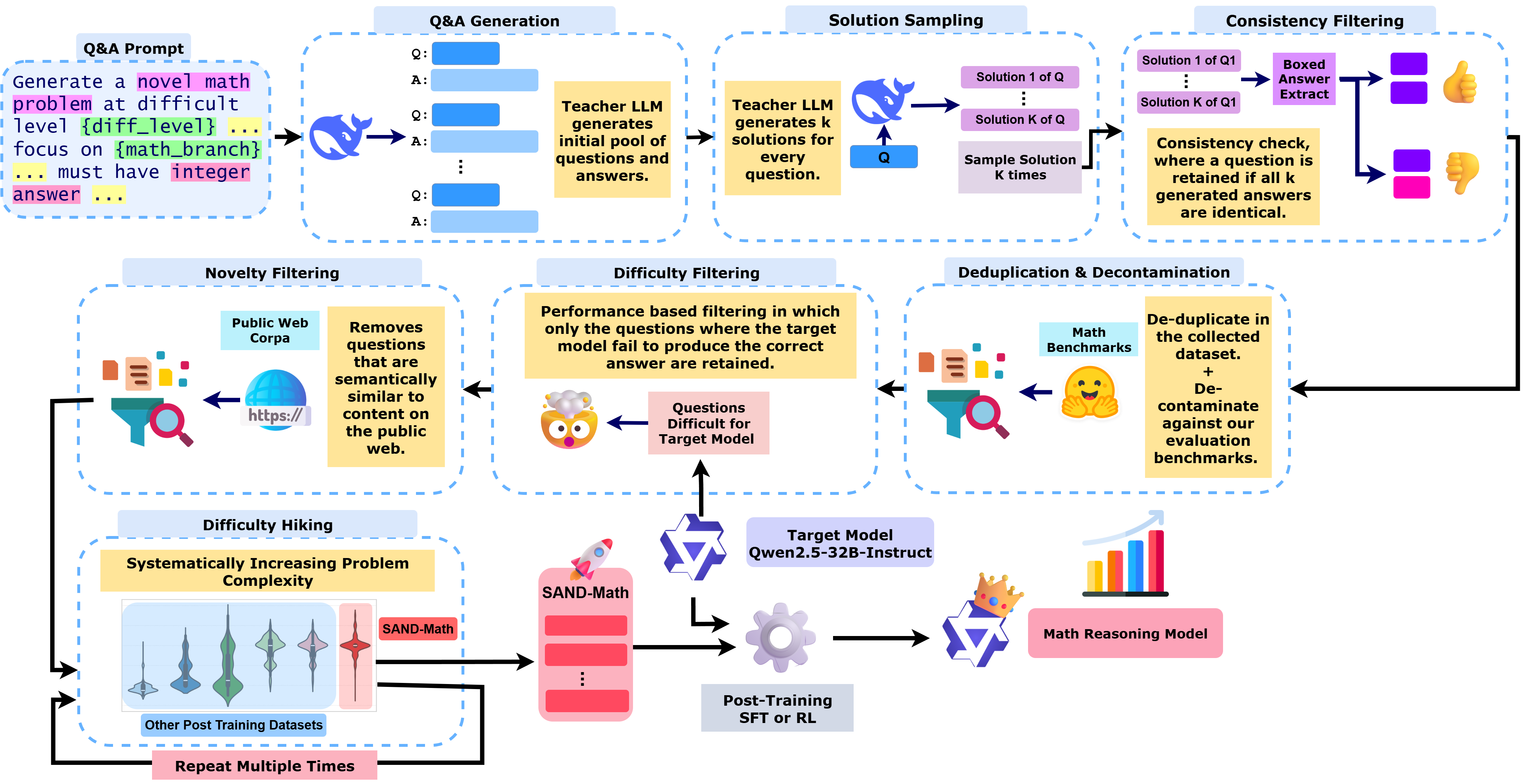}
  \caption{Data Generation and Filtering pipeline for SAND-Math.} 
  
  \label{fig:pipeline}
\end{figure*}
\section{Introduction}  
\label{sec:introduction}

Mathematical problem-solving is emerging as a pivotal area of interest as the frontier of Large Language Model (LLM) capabilities rapidly advances into domains that demand sophisticated reasoning. This critical capability is essential for driving innovation and efficiency across numerous industries, like quantitative finance, scientific research, and engineering. State-of-the-art (SOTA) reasoning models such as DeepSeek-R1 \citep{deepseekai2025deepseekr1incentivizingreasoningcapability}, Exaone \citep{research2024exaone35serieslarge}, OpenAI o3 \citep{openai2024gpto3}, and Gemini 2.5 Pro \citep{comanici2025gemini25pushingfrontier} have demonstrated remarkable performance on challenging mathematical and reasoning benchmarks \citep{he2024olympiadbench,aime2024,hendrycksmath2021,cobbe2021gsm8k,pmlr-v267-lin25i,rein2023gpqagraduatelevelgoogleproofqa,mishra2025tttbenchbenchmarkevaluatingreasoning}. However, these proprietary models typically do not disclose their training data or generation methods, creating a major barrier for broader adoption. Recent work such as LIMO \citep{ye2025limoreasoning} and S1 \citep{muennighoff2025s1simpletesttimescaling} reveals that reasoning depends more on data quality—especially highly challenging problems—than on sheer scale. Yet, sourcing such problems is a bottleneck; datasets like NuminaMath \citep{numina_math_datasets} and OpenR1 \citep{openr1} rely on manual curation from olympiyads \& competitions, which is labor-intensive and limited, making synthetic data generation a necessary direction.

However, current synthetic methods, including KPDDS \citep{huang2024keypointdrivendatasynthesisenhancement}, MetaMathQA \citep{yu2023metamath}, WizardMath \citep{luo2025wizardmathempoweringmathematicalreasoning} and OpenMathInstruct2 \citep{toshniwal2024openmath2}, primarily remix existing train set from benchmarks like GSM8K \citep{cobbe2021gsm8k} and MATH \citep{hendrycksmath2021}, limiting their ability to surpass seed data difficulty. While MATH$^2$ \citep{shah2025aiassistedgenerationdifficultmath} proposed a human-in-the-loop approach for harder problems, it remains difficult to scale. We hypothesize that SOTA LLMs possess \textbf{metacognitive abilities} \citep{didolkar2024metacognitivecapabilitiesllmsexploration}, enabling them to implicitly model the characteristics of difficult math problems and generate new problems from minimal prompts. We leverage this to build a fully automated pipeline~\ref{fig:pipeline} that generates, filters, and difficulty-hikes math problems, while ensuring correctness, novelty, and escalating complexity. As shown in Figure \ref{fig:dataset_diff_ratings}, our dataset surpasses prior synthetic work in difficulty and rivals human curated Olympiad level problems.

The main contributions of this work are as follows:
\begin{itemize}
    \item We introduce \textbf{SAND-Math}, a novel pipeline for generating challenging synthetic math problems using the hypothesized metacognitive abilities of SOTA models, without relying on any seed dataset.
    \item Our proposed pipeline ensures rigorous controls for correctness, novelty, and a systematic \textbf{difficulty hiking} mechanism.
    \item We provide a new high-difficulty synthetic dataset and a fine-tuned model that demonstrate significant improvements in mathematical reasoning.
\end{itemize}

\section{SAND-Math Pipeline}
\label{snad-math-pipeline}

We present an end-to-end synthetic data generation and filtering pipeline that produces sufficiently challenging mathematics problems and solutions designed to elicit LLM reasoning. The pipeline proceeds through multiple stages, culminating in a novel \textit{Difficulty Hiking} method that systematically elevates problem complexity.

\subsection{Question and Solution Generation}
\label{subsec:question_and_solution_generation}
We begin by generating an initial set of questions with our teacher LLM, $\mathcal{M}_{\text{teacher}}$, using an empirically optimized prompt (Appendix~\ref{app:question_prompt}). To keep the LLM grounded to generate solvable questions, $\mathcal{M}_{\text{teacher}}$ is prompted to co-generate a corresponding solutions, though these are not the final solutions that are used for subsequent training. For each generated question $q_i$, we then prompt $\mathcal{M}_{\text{teacher}}$ a second time (Appendix~\ref{app:solution_prompt}) to generate $k$ distinct solutions. This results in a set of reasoning traces $\mathcal{T}_i = \{ (r_{ij}, a'_{ij}) \}_{j=1}^k$, where $r_{ij}$ is the step-by-step reasoning and $a'_{ij}$ is the final answer. This process yields our initial dataset $\mathcal{D}_1 = \{ (q_i, \mathcal{T}_i) \}_{i=1}^{23,437}$, comprising 23,437 question and solution pair sets.

\subsection{Solution Correctness Filtering}
\label{subsec:solution_correctness_filtering}
Next, we apply self-consistency \citep{wang2023selfconsistencyimproveschainthought} to $\mathcal{D}_1$: a question is retained only if all $k$ answers agree ($a'_{i1} = a'_{i2} = \dots = a'_{ik}$). One reasoning trace–answer pair $(r_i, a_i)$ is sampled from $k$ pairs to form $\mathcal{D}_{cons}$, yielding 17,578 consistent examples ($\sim$74\% of original yield).
\subsection{De-duplication and Decontamination}
\label{subsec:decontamination_deduplication}
We ensure novelty and integrity of $\mathcal{D}_{cons}$ via two steps. First, semantic de-duplication is performed using the \texttt{semhash} framework \citep{minishlab2025semhash} with \texttt{minisilab/potion-base-8M} \citep{minishlab2024model2vec}, removing 1,293 duplicates (7.3\%) at a 0.99 similarity threshold, chosen through sensitivity analysis to preserve diversity yet removing duplicates. For the full analysis of threshold see Appendix~\ref{app:sensitivity_analysis}. Next, to prevent test set leakage, we use a two-stage decontamination pipeline: an efficient retrieval model finds the top-5 candidates from test benchmarks, which are then semantically verified by our judge model, $\mathcal{M}_{\text{judge}}$, guided by the prompt (Appendix~\ref{app:decontamination_prompt}). This hybrid design is optimized for scalability, with our model choice empirically justified in Appendix~\ref{app:decontamination_model_choice}, and removed only 4 questions, leaving 16,281 examples.
\subsection{Difficulty Filtering and Rating}  
\label{subsec:difficulty_filtering}
At this stage, the dataset undergoes two parallel processes for difficulty curation. First, for performance-based filtering, we use a solver model $\mathcal{M}_{\text{solver}}$, and retain only the questions it answers incorrectly. This step ensures empirical difficulty, yielding $\mathcal{D}_{\text{diff}}$ with 9,211 problems (a 56.6\% retention rate). Concurrently, our judge model $\mathcal{M}_{\text{judge}}$, assigns a fine-grained 1-10 difficulty rating to each question, guided by a rubric and reference examples from AoPS\footnote{\url{https://artofproblemsolving.com/wiki/index.php/AIME_Problems_and_Solutions}} (see Appendix~\ref{app:diffclass_prompt}). These ratings are crucial for dataset analysis and as direct input for our Difficulty Hiking module.

\subsection{Novelty Filtering}
To ensure our dataset can be safely combined with public corpora without redundancy, we apply a final novelty filter to $\mathcal{D}_{\text{diff}}$. Each question is used as a web search query, and its maximum semantic similarity to the top-10 results is computed with strong sentence embedding model, following Algorithm~\ref{alg:novelty_check}. Any question with similarity above $\tau=0.85$ is discarded. This removed 4\% of the data, producing our final dataset of 8,842 novel problems, which we call \textbf{SAND-Math}.  

\begin{wrapfigure}{r}{0.5\textwidth}
    \begin{center}
    \vspace{-1em}
        \includegraphics[width=0.48\textwidth]{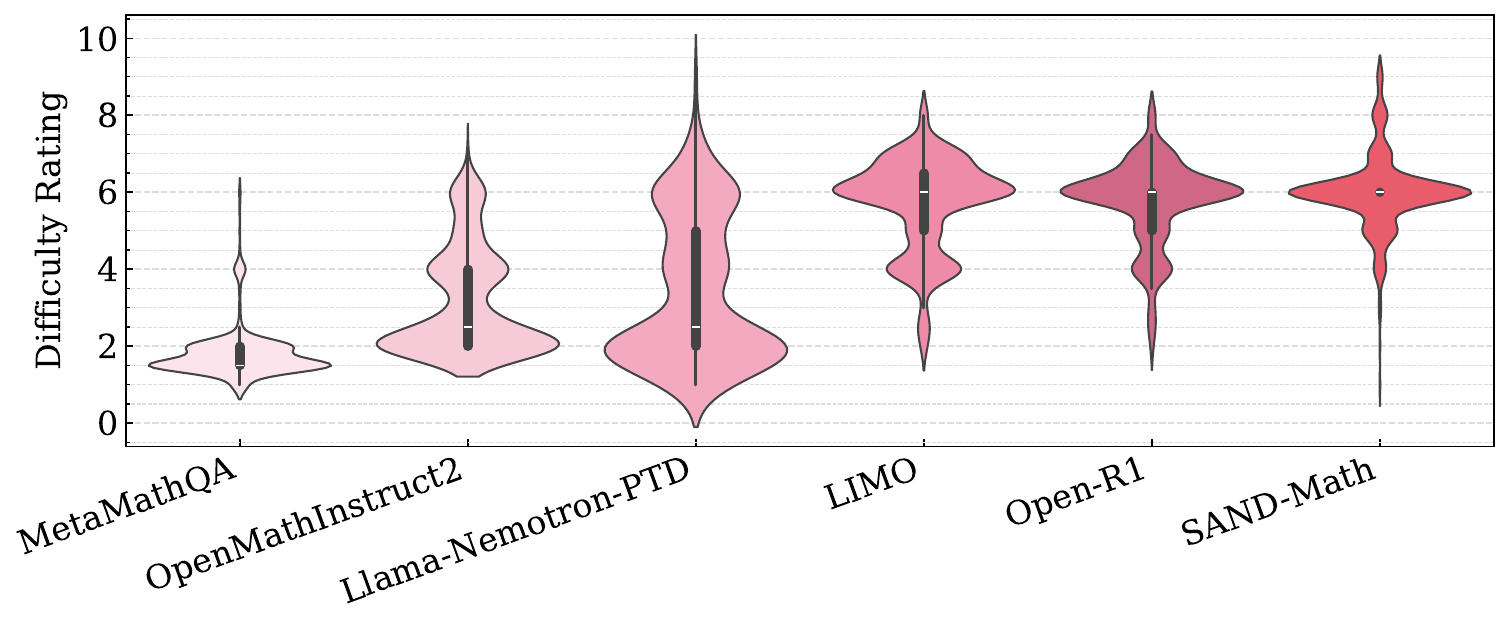}
      \end{center}
      \caption{Difficulty distribution of SAND-Math (500) dataset compared with other math datasets}
      \vspace{-1em}
    \label{fig:dataset_diff_ratings}
\end{wrapfigure}
\subsection{Difficulty Hiking: Systematically Increasing Problem Complexity}

To further raise the difficulty of our initially generated questions, we introduce \textbf{Difficulty Hiking}. This method involves re-prompting the teacher model $\mathcal{M}_{\text{teacher}}$ to rewrite a question, guided by its original version, its difficulty rating, and a forced synthesis of a same-branch theorem with a cross-domain concept (see Appendices~\ref{app:difficulty_hiking_prompt}, \ref{app:hiking_example}, \ref{app:math_tax} for full details). In a single iteration, this process proved highly effective: it not only shifted the mean difficulty rating from 5.02 to 5.98 (Figure~\ref{fig:diffhike_disirbution_bar}) but, more importantly, finetuning on this enhanced data boosted downstream performance from 72.94 to \textbf{74.39} (Table~\ref{tab:diffhike_results}). While this process is iterative, we only show results with a single hiking step in this work.

\section{Experiments and Results}  
  
\paragraph{Implementation Details}
Our pipeline assigned distinct roles to specific models to ensure reproducibility. Our teacher model, $\mathcal{M}_{\text{teacher}}$, which performed initial question generation (temp=0.8), solution generation with $k=2$ samples (temp=0.6), and question modification during the Difficulty Hiking stage, was DeepSeek-R1~\citep{deepseekai2025deepseekr1incentivizingreasoningcapability}. All judging and rating tasks were conducted by our judge model, $\mathcal{M}_{\text{judge}}$, which was Llama-3.3-70B-Instruct~\citep{grattafiori2024llama3herdmodels}, including verifying solution consistency, acting as the decontamination judge, and assigning fine-grained difficulty scores (averaged over 3 runs for stability). The solver model for performance-based difficulty filtering, $\mathcal{M}_{\text{solver}}$, was \texttt{Qwen2.5-32B-Instruct} \citep{qwen2.5}. Specialized filtering employed semhash~\citep{minishlab2025semhash} with a 0.99 similarity threshold for de-duplication, and novelty checks used \texttt{Alibaba-NLP/gte-Qwen2-7B-instruct} embeddings \cite{li2023towards}  via a self-hosted SearXNG~\citep{searxng_github}
instance accessed through LangChain. The entire pipeline was executed on a single node with 8$\times$~AMD Instinct\textsuperscript{TM} MI300X GPUs, with throughput enhanced using optimizations described in the Supercharge DeepSeek-R1 ROCm\textsuperscript{TM} blog post~\citep{deepseekinferenceonamd}.

All models were finetuned on Qwen2.5-32B-Instruct~\citep{qwen2.5} using the LLaMA-Factory~\citep{zheng2024llamafactory} framework within the \texttt{rocm/pytorch-training} Docker container~\citep{rocmpytorchtrainingdocker}, running on a single node with 8$\times$~AMD Instinct\textsuperscript{TM} MI300X GPUs. We employed DeepSpeed ZeRO-3~\citep{rajbhandari2020zeromemoryoptimizationstraining} for memory efficiency, adopting the LIMO setup~\citep{ye2025limoreasoning} with full-parameter supervised finetuning, a learning rate of 5e-6, and 10 training epochs (details in Appendix~\ref{app:training_details}). To contextualize SAND-Math, we finetuned against equal-sized splits of open-source synthetic datasets (OpenMathInstruct-v2~\citep{toshniwal2024openmath2}, MetamathQA~\citep{yu2023metamath}) and high-difficulty real-world corpora (OpenR1-Math-220k~\citep{openr1}, Llama-Nemotron-Post-Training-Dataset~\citep{bercovich2025llamanemotronefficientreasoningmodels}). Evaluation was performed using the LIMO codebase~\citep{ye2025limoreasoning} on AIME (2024/2025), AMC, and MATH500. We report \texttt{pass@1} with 16 samples per problem ($n=16$, temp=0.7) for AIME/AMC, and greedy decoding accuracy (temp=0) for MATH500.

\paragraph{Experimental Design}
To evaluate SAND-Math, we adopt a two-pronged approach reflected in Table~\ref{tab:main_results_updated}. First, we assess its standalone quality by finetuning a model on a 500-sample subset, comparing it directly against subsets of other prominent real-world and synthetic datasets. Second, and more critically, we evaluate its effectiveness in a practical, real-world scenario: data augmentation. High-quality synthetic data is rarely used in isolation; its primary utility lies in enhancing strong, existing datasets. For this purpose, we selected the LIMO dataset as our baseline for augmentation. LIMO is a high-quality, human-curated corpus that is potent yet small enough (817 examples) that the impact of adding 500 new samples is not diluted, allowing for a clear assessment of each dataset's contribution.

\paragraph{Results}


\begin{table*}[t!]
\centering
\caption{\textbf{Post-training performance of Qwen2.5-32B trained with SAND-Math vs. other datasets}, for 10 epochs. We report pass@1 accuracy (\%). SAND-Math achieves competitive standalone results and yields the largest gains when augmenting the LIMO baseline.
}
\label{tab:main_results_updated}

\resizebox{0.9\textwidth}{!}{
\begin{tabular}{l
                S[table-format=4.2, detect-weight] 
                S[table-format=2.2, detect-weight] S[table-format=2.2, detect-weight] S[table-format=2.2, detect-weight]
                S[table-format=2.2, detect-weight] S[table-format=2.4, detect-weight]}
\toprule
\textbf{Training Data Configuration} & {\textbf{Data Sample Size}} & {\textbf{AIME25}} & {\textbf{AIME24}} & {\textbf{AMC}} & {\textbf{MATH-500}} & {\textbf{Average}} \\
\midrule
\multicolumn{7}{c}{\textit{\textbf{Setting 1: Baselines}}} \\
Qwen 2.5 32B     & {-}   & 13.33 & 16.67 & 64.38 & 83.40 & 44.44 \\
LIMO (SFT baseline)           & 817   & 44.50 & 56.30 & 91.41 & 93.80 & 71.50 \\
\midrule
\multicolumn{7}{c}{\textit{\textbf{Setting 2: Standalone Finetuning (vs. Real-World Problems Datasets)}}} \\
\textbf{SAND-Math (ours)}     & 500   & 41.88 & 52.08 & 89.84 & 92.60 & 69.10 \\
openr1\_math                  & 500   & \bfseries 43.96 & \bfseries 53.12 & \bfseries 90.62 & \bfseries 93.40 & \bfseries 70.27 \\
llama\_nemotron               & 500   & 17.50 & 20.62 & 65.62 & 84.80 & 47.13 \\
\midrule
\multicolumn{7}{c}{\textit{\textbf{Setting 3: Augmentation Performance (vs. Real-World Problems Datasets)}}} \\
\textbf{LIMO + SAND-Math (ours)}   & {817 + 500} & \bfseries 48.89 & \bfseries 57.92 & \bfseries 92.50 & \bfseries 94.00 & \bfseries 73.32 \\
LIMO + openr1\_math                & {817 + 500} & 47.71 & 56.04 & \bfseries 92.50 & 93.80 & 72.51 \\
LIMO + llama\_nemotron             & {817 + 500} & 26.04 & 36.46 & 71.56 & 86.20 & 55.06 \\
\midrule
\multicolumn{7}{c}{\textit{\textbf{Setting 4: Augmentation Performance (vs. Synthetic Datasets)}}} \\
\textbf{LIMO + SAND-Math (ours)}   & {817 + 500} & \bfseries 48.89  & \bfseries 57.92 & \bfseries 92.50 & \bfseries 94.00 & \bfseries 73.32 \\
LIMO + MetamathQA                & {817 + 500} & 31.04 & 46.25 & 47.24 & 56.40 & 45.23 \\
LIMO + OpenmathInstruct          & {817 + 500} & 18.13 & 38.96 & 64.53 & 72.40 & 48.50 \\
\bottomrule
\end{tabular}
}

\end{table*}

Our experiments, summarized in Table~\ref{tab:main_results_updated}, validate SAND-Math as a high-quality data source through both standalone and augmentation evaluations.

In the standalone setting (Setting 2 from the Table~\ref{tab:main_results_updated}), finetuning on 500 samples of SAND-Math achieves an average score of 69.10. This result is highly competitive with the 70.27 score from \texttt{openr1\_math}, a high-quality human-curated dataset, demonstrating that our fully automated pipeline produces data of comparable quality to expert-created corpora.

However, the primary strength of SAND-Math is revealed in the data augmentation setting (Setting 3 \& 4 in  Table~\ref{tab:main_results_updated}). When used to augment the strong LIMO baseline, just 500 samples of SAND-Math yield a top average score of 73.32. This surpasses the performance of augmenting LIMO with \texttt{openr1\_math} (72.51) and significantly outperforms augmentation with other synthetic datasets like \texttt{MetamathQA} and \texttt{OpenMathInstruct}. This result confirms our main hypothesis: SAND-Math's targeted difficulty provides a unique and data-efficient signal that effectively enhances strong foundation models.

\begin{table*}[!t]  
  \caption{ \textbf{Impact of Difficulty Hiking on Finetuning Performance.} Ablation study showing the pass@1 accuracy (\%) of the \textbf{Qwen2.5-32B} model. Augmenting the LIMO baseline with our standard Difficulty Hiked data (DH) yields the best average performance over the non-hiked Base data. We also ablate the effect of applying a length filter to remove verbose examples (DH\_w\_LF).}
  \label{tab:diffhike_results}  
  \centering 
  \resizebox{0.8\textwidth}{!}{
      \begin{tabular}{lcrrrrr}  
        \toprule  
        \textbf{Dataset}& \textbf{Data Size} & \textbf{AIME25} &  \textbf{AIME24} & \textbf{AMC24} & \textbf{MATH500} & \textbf{Average}\\  
        \midrule  
        \textit{LIMO only (Baseline)} & \textit{817} & \textit{44.5} & \textit{56.3} & \textit{91.41} & \textit{93.8} & \textit{71.50} \\
        \midrule
        LIMO + SAND-Math (Base) & 817 + 1500 & 46.38 & 59.09 & 92.71 & 93.6 & 72.94 \\  
        LIMO + SAND-Math (DH) & 817 + 1500 & \textbf{49.23} & 60.55 & 93.17 & \textbf{94.6} & \textbf{74.39} \\  
        LIMO + SAND-Math (DH\_w\_LF)  & 817 + 1500 & \textbf{49.23}  & \textbf{60.83} & \textbf{93.28} & 93 & 74.08 \\  
        \bottomrule  
      \end{tabular}  
      }
\end{table*}

The key to this strong performance is the dataset's elevated problem complexity, which validates our central hypothesis. As shown in Figure~\ref{fig:dataset_diff_ratings}, SAND-Math is centered at a much higher mean difficulty than competing datasets. This principle is best exemplified by our \textbf{Difficulty Hiking} methodology. The ablation in Table~\ref{tab:diffhike_results} shows its direct impact: finetuning on difficulty-hiked data (\textbf{DH}) boosts the average score from \textbf{72.94$\rightarrow$74.39}, a significant gain over the non-hiked \textbf{Base} data. To investigate the effect of potentially verbose outputs from the hiking process, we also test a variant, \textbf{DH\_w\_LF}, where we apply a length filter to the DH data. This version scores slightly lower at 74.08, suggesting that even the longer, more detailed reasoning chains generated during hiking are beneficial for model training. This highlights the exceptional sample efficiency of our approach, where adding just 1500 enhanced examples to the strong LIMO baseline yields substantial improvements. This occurs because the hiking process transforms the dataset's composition by increasing the density of highly complex problems (Figure~\ref{fig:diffhike_disirbution_bar}), creating a more rigorous training curriculum that directly improves the model's reasoning capabilities.

To validate that these accuracy gains reflect true improvements in reasoning, we conducted further qualitative and failure analyses (detailed in Appendix~\ref{app:reasoning_analysis}). A large-scale evaluation confirms our model produces solutions with superior logical soundness and clarity. Furthermore, a failure analysis reveals that the model's remaining errors are concentrated in complex logical deduction on the hardest problems, validating our paper's central thesis. Finally, our difficulty-hiked data also demonstrates superior data efficiency, yielding greater performance gains with fewer examples, as shown by its almost linear scaling law as shown in Appendix~\ref{app:scaling_analysis}.

\section{Conclusion}  
This work addresses the scarcity of high-quality data for mathematical LLMs by introducing \textbf{SAND-Math}, an iterative pipeline that generates novel, complex problems. Our integrated \textit{Difficulty Hiking} method further elevates problem complexity, yielding significant performance gains even for smaller models without having to rely on human mathemtical experts for curating such datasets. Beyond mathematics, this approach offers a scalable paradigm for generating high-quality, reasoning-focused datasets, opening the door to advancing LLM capabilities across multiple scientific domains.  

\section*{Limitations}  
Our study demonstrates the effectiveness of the SAND-Math pipeline, but we also acknowledge few considerations regarding its implementation and scope. The output data quality and correctness and thus also the accuracy of downstream student model are directly dependent on the capabilities of the teacher models. The current pipeline with DeepSeek-R1 resulted in just a 41.9\% yield. However, this is not a fundamental ceiling. As shown in an ablation study in Appendix~\ref{app:yield_ablation} (Table~\ref{tab:yield_comparison}), using the more recent GPT-OSS 120B model \citep{openai2025gptoss120bgptoss20bmodel} increased this yield dramatically to 74.0\% and also producing higher difficulty level questions while being almost 5.2X faster. This confirms a clear and promising path to greater scalability and efficiency as foundation models continue to improve. Finally, our finetuning experiments were conducted on a small representative sample of the full SAND-Math dataset to serve as a proof-of-concept under given compute budget. A large-scale post-training is a valuable next step to see how much further we can improve through this fully automated data generation approach.

\bibliographystyle{unsrtnat}
\bibliography{custom}

\appendix
\section{Pipeline Yield and Teacher Model Ablation}\label{app:yield_ablation}

As discussed in the Limitations section, the yield of the SAND-Math pipeline is directly dependent on the capabilities of the teacher model. To demonstrate the scalability and future potential of our approach, we conducted an ablation study comparing our in-paper teacher model, DeepSeek-R1, with a more recent, powerful model. Table~\ref{tab:yield_comparison} shows that using GPT-OSS 120B as the teacher model nearly doubles the combined data yield from 41.9
\begin{table}[h]
\caption{\textbf{Pipeline yield is dependent on teacher model quality.} A more capable teacher model nearly doubles the data yield, demonstrating the pipeline's scalability.}
\label{tab:yield_comparison}
\centering
\begin{tabular}{lccc}
\toprule
\textbf{Teacher Model} & \textbf{Self-Consistency Rate} & \textbf{Difficulty Filter Rate} & \textbf{Combined Yield} \\
\midrule
DeepSeek-R1 (in paper) & 74.0\% & 56.6\% & 41.9\% \\
\textbf{GPT-OSS 120B (new)} & \textbf{94.0\%} & \textbf{78.8\%} & \textbf{74.0\%} \\
\bottomrule
\end{tabular}

\end{table}

\section{Detailed Analysis of Difficulty Hiking}
\label{app:difficulty_hiking_analysis}

\begin{figure}[h]
    \centering
    \includegraphics[width=\textwidth]{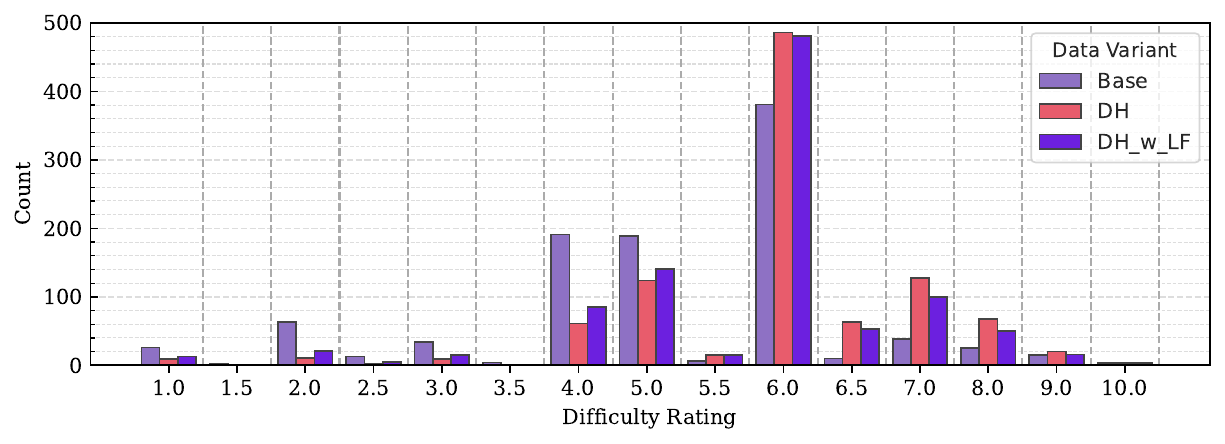}
    
    \caption{\textbf{Impact of Difficulty Hiking on Data Distribution.} Comparison of difficulty ratings for a SAND-Math sample across three variants: the original Base (non-difficulty hiked) data, the DH (Difficulty Hiked) version, and the DH\_w\_LF version (DH with a 32k length filter). The hiking process successfully shifts the questions difficulty distribution more towards (6 to 8) range.}
    
    \label{fig:diffhike_disirbution_bar}
\end{figure}

To better illustrate the effectiveness of our Difficulty Hiking method, Figure~\ref{fig:diffhike_disirbution_bar} provides a visual comparison of the difficulty distribution for a sample of SAND-Math data. The plot shows the original distribution (\texttt{Base}), the distribution after applying Difficulty Hiking (\texttt{DH}), and an ablation with an additional length filter (\texttt{DH\_w\_LF}). The visualization clearly shows how the process transforms questions from the mid-difficulty range (e.g., ratings 4.0-5.0) into the more challenging 6.0-8.0 range, which is critical for improving the reasoning capabilities of the student model.

\section{Deeper Analysis of Reasoning and Model Failures}
\label{app:reasoning_analysis}
To move beyond accuracy-based metrics, we conducted further analyses to evaluate the quality of the model's reasoning process and understand its remaining failure modes.

\paragraph{Qualitative Analysis of Solution Quality.}
We performed a large-scale qualitative evaluation on 480 generated solutions. An impartial LLM judge scored solutions from our baseline and Difficulty Hiked models on a 1-5 scale across four aspects of reasoning quality. As shown in Table~\ref{tab:qualitative_analysis}, our model demonstrated consistent improvements across all dimensions, indicating not just better final answers, but higher-quality mathematical reasoning.

\begin{table}[h]
\centering
\caption{\textbf{Qualitative evaluation of reasoning quality.} Our SAND-Math trained model shows superior performance across multiple dimensions of mathematical reasoning beyond simple accuracy.}
\label{tab:qualitative_analysis}
\resizebox{\columnwidth}{!}{
\begin{tabular}{lccc}
\toprule
\textbf{Metric (1-5 Scale)} & \textbf{Baseline (LIMO)} & \textbf{LIMO + SAND-Math (DH)} & \textbf{Improvement} \\
\midrule
Correctness (Logical Soundness) & 3.03 & 3.20 & +0.17 \\
Faithfulness (Avoiding Hallucination) & 3.54 & 3.64 & +0.10 \\
Clarity (Readability of Steps) & 3.03 & 3.10 & +0.07 \\
Conciseness (Solution Efficiency) & 2.06 & 2.18 & +0.12 \\
\bottomrule
\end{tabular}
}
\end{table}

\paragraph{Failure Analysis.}
We also analyzed our best model's errors on the AIME 2025 benchmark. As shown in Table~\ref{tab:failure_analysis}, the dominant failure mode is a breakdown in the core logical deduction process (35.8\%), especially on the most difficult problems. This finding validates our paper's central thesis: the primary bottleneck is complex reasoning, and the most effective solution is to train on more problems at an elevated difficulty, which our pipeline is designed to generate at scale.

\begin{table}
\centering
\caption{\textbf{Distribution of error types} for our best model on AIME 2025. The prevalence of reasoning errors highlights the need for more complex training data.}
\label{tab:failure_analysis}
\begin{tabular}{lc}
\toprule
\textbf{Error Type} & \textbf{Percentage of Failures} \\
\midrule
Reasoning Error (Logical breakdown) & 35.8\% \\
Misinterpretation (Problem setup) & 11.9\% \\
Calculation Error (Arithmetic) & 5.2\% \\
Hallucination Error (Irrelevant steps) & 4.6\% \\
\bottomrule
\end{tabular}
\end{table}

\section{Data Efficiency and Scaling Analysis}
\label{app:scaling_analysis}
\begin{figure}[h!]
    \centering
    \includegraphics[width=0.7\columnwidth]{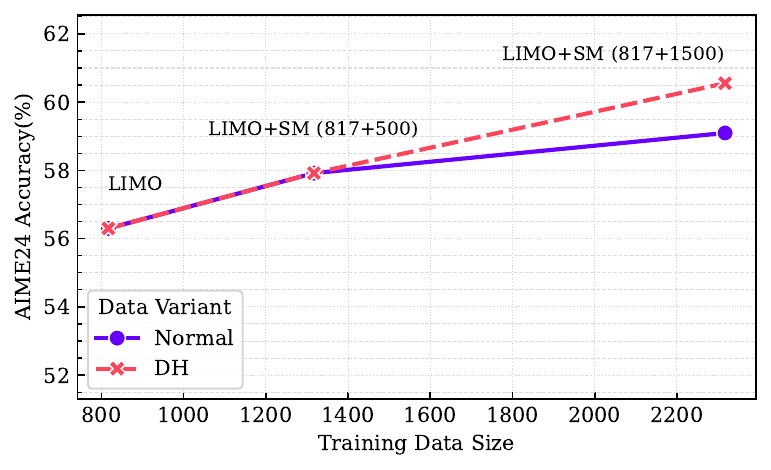}
    \caption{Performance trend when augmenting the LIMO training data with SAND-Math (SM) samples. The 'DH' (Difficulty Hiked) condition shows greater improvement with 1500 additional samples.}
    \label{fig:finetuning_trend}
\end{figure}

To analyze the data efficiency of our approach, we incrementally added samples to the LIMO baseline and plotted the performance on the AIME24 benchmark. As shown in Figure~\ref{fig:finetuning_trend}, while both base ('Normal') and difficulty-hiked ('DH') versions of SAND-Math improve performance, the 'DH' data exhibits a much steeper improvement trajectory. Augmenting with just 1500 'DH' samples significantly outperforms the base data augmentation. This demonstrates that for a given data budget, difficulty-hiked data provides a superior return on investment, enabling greater performance gains with fewer examples.

\section{Sensitivity Analysis for De-duplication Threshold}
\label{app:sensitivity_analysis}

To empirically validate our choice of the 0.99 cosine similarity threshold for internal de-duplication (Section~\ref{snad-math-pipeline}), we performed a sensitivity analysis. The goal was to find the optimal balance between aggressively removing potential duplicates and precisely identifying only true semantic duplicates, thereby preserving the dataset's diversity. We tested three thresholds (0.99, 0.97, and 0.95), measuring both the percentage of data flagged for removal and the precision of this flagging, which was verified by an LLM judge (\texttt{o4-mini}) on a sample of filtered pairs.

The results, presented in Table~\ref{tab:dedup_sensitivity}, highlight a critical trade-off. While a lower, more lenient threshold like 0.95 flags a substantial portion of the data (44.5\%), its precision is exceptionally low (24.0\%), indicating that it would incorrectly remove a large volume of unique, non-redundant problems. In contrast, the 0.99 threshold offers a much higher precision of 52.0\%, ensuring that the flagged content is highly likely to be a true duplicate. Therefore, we selected 0.99 as the most methodologically sound choice, as it effectively removes near-duplicates while minimizing the risk of over-filtering and preserving the valuable diversity of our generated dataset.

\begin{table}
    \centering
    \caption{Sensitivity analysis for the semantic de-duplication threshold. The 0.99 threshold provides the best balance, maximizing precision while effectively removing near-duplicates.}
    \label{tab:dedup_sensitivity}
    \resizebox{\columnwidth}{!}{
    \begin{tabular}{lccl}
        \toprule
        \textbf{Cosine Sim. Threshold} & \textbf{\% Data Flagged} & \textbf{Precision (\% True Duplicates)} & \textbf{Outcome} \\
        \midrule
        \textbf{0.99 (Our Choice)} & \textbf{7.4\%} & \textbf{52.0\%} & \textbf{Effective balance} \\
        0.97 & 26.7\% & 38.0\% & Aggressive filtering \\
        0.95 & 44.5\% & 24.0\% & High loss of unique data \\
        \bottomrule
    \end{tabular}%
    }
\end{table}

\section{Justification for Decontamination Model Choice}
\label{app:decontamination_model_choice}

Our decontamination pipeline (Section~\ref{subsec:decontamination_deduplication}) employs a two-stage design: a fast sentence-transformer model for candidate retrieval, followed by a powerful LLM (\texttt{Llama-3.3-70B-Instruct} \citep{grattafiori2024llama3herdmodels}) for precise judging. This design requires the retrieval model to be extremely efficient while still being effective at identifying the strongest potential duplicates. We selected \texttt{paraphrase-multilingual-MiniLM-L12-v2} \citep{reimers-2019-sentence-bert} for this role over larger, more recent models like \texttt{gte-Qwen2-1.5B} \citep{li2023towards}. This choice was based on an empirical comparison of both efficiency and effectiveness.

\paragraph{Efficiency.}
For large-scale data processing, retrieval speed is critical. Our chosen \texttt{MiniLM} model provides an order-of-magnitude improvement in encoding speed compared to the larger alternative, as shown in Table~\ref{tab:decontam_efficiency}. This makes it far more suitable for a scalable data generation pipeline.

\begin{table}
\centering
\caption{Efficiency comparison of sentence transformer models for candidate retrieval. Our chosen model is over 13 times faster.}
\label{tab:decontam_efficiency}
\begin{tabular}{lcc}
\toprule
\textbf{Model} & \textbf{Time to Encode 10k Samples} & \textbf{Speed Advantage} \\
\midrule
\textbf{paraphrase-MiniLM (Ours)} & \textbf{4.94s} & \textbf{13.5$\times$ faster} \\
gte-Qwen2-1.5B (Larger Alternative) & 66.54s & - \\
\bottomrule
\end{tabular}%
\end{table}

\paragraph{Effectiveness.}
While speed is important, the retrieval model must also be effective at its core task: surfacing the most likely duplicates for the LLM judge to review. To test this, we conducted a targeted experiment. For each retrieval model, we identified the 30 pairs it was most confident were duplicates (i.e., those with the highest cosine similarity). We then used a frontier LLM judge to score the true semantic similarity of these high-confidence pairs. As shown in Table~\ref{tab:decontam_effectiveness}, the candidates surfaced by our faster \texttt{MiniLM} model were judged to be equally, if not more, semantically similar.

\begin{table}
\centering
\caption{Effectiveness comparison. We report the average similarity score assigned by a frontier LLM judge to the top 30 most confident pairs retrieved by each model. Higher scores indicate better candidate retrieval.}
\label{tab:decontam_effectiveness}
\begin{tabular}{lc}
\toprule
\textbf{Model} & \textbf{Avg. LLM Similarity Score (Top 30 Pairs)} \\
\midrule
\textbf{paraphrase-MiniLM (Ours)} & \textbf{0.41} \\
gte-Qwen2-1.5B (Larger Alternative) & 0.36 \\
\bottomrule
\end{tabular}%
\end{table}

This analysis confirms that our choice of \texttt{paraphrase-multilingual-MiniLM-L12-v2} is optimal for its role in our pipeline, delivering a massive improvement in efficiency with no compromise on the effectiveness of the decontamination process.

\section{Detailed Training Steps}
\label{app:training_details}

All models trained in this work are full-parameter finetuned using the \texttt{LLaMA-Factory}~\citep{zheng2024llamafactory} framework. Training was conducted on a single node equipped with 8$\times$~AMD Instinct\textsuperscript{TM} MI300X GPUs. This section provides the key hyperparameters and a brief guide to the training setup on AMD GPUs.

\subsection{Hyperparameters}
The core hyperparameters used for our full-parameter finetuning experiments are detailed in Table~\ref{tab:hyperparams}.

\begin{table}[h!]
\caption{Key training hyperparameters used in our experiments.}
\label{tab:hyperparams}
\centering
\begin{tabular}{@{}ll@{}}
\toprule
\textbf{Hyperparameter}             & \textbf{Value}         \\ \midrule
Learning Rate                       & \texttt{5.0$e$-6}        \\
LR Scheduler Type                   & \texttt{cosine}        \\
Warmup Ratio                        & \texttt{0.0}           \\
Number of Training Epochs           & \texttt{10}            \\
Gradient Accumulation Steps         & \texttt{1}             \\
Cutoff Length                       & \texttt{32,768}         \\
Flash Attention Implementation      & \texttt{fa2}           \\
DeepSpeed Strategy                  & \texttt{ZeRO-3}        \\ \bottomrule
\end{tabular}

\end{table}

\subsection{Training Setup}
The following steps outline the process for replicating our training environment.

\begin{enumerate}
    \item \textbf{Launch the Docker Container:}
    The experiments were run inside the ROCm\textsuperscript{TM} PyTorch training container~\citep{rocmpytorchtrainingdocker}.
\begin{lstlisting}[language=bash]
docker run -it \
    --ipc=host \
    --cap-add=SYS_PTRACE \
    --network=host \
    --device=/dev/kfd \
    --device=/dev/dri \
    --security-opt \
    seccomp=unconfined \
    --group-add video \
    --privileged -w /workspace \
    rocm/pytorch-training:v25.6
\end{lstlisting}

    \item \textbf{Install LLaMA-Factory:}
    Clone the repository and install the required dependencies. A specific version of DeepSpeed was used for compatibility.
\begin{lstlisting}[language=bash]
git clone --depth 1 \
    https://github.com/hiyouga/LLaMA-Factory.git
cd LLaMA-Factory
pip install -e ".[torch,metrics]" --no-build-isolation
pip install deepspeed==0.16.9
\end{lstlisting}

    \item \textbf{Prepare Training Files:}
    \begin{itemize}
        \item Download the required SAND-Math training splits from HuggingFace and format them into the Alpaca-style JSON format, as described in the documentation\footnote{\href{https://github.com/hiyouga/LLaMA-Factory/tree/main/data}{\url{https://github.com/hiyouga/LLaMA-Factory/tree/main/data}}}.        
        \item Add a corresponding entry for the new dataset file in \texttt{data/dataset\_info.json}.
        \item Create a training configuration YAML file. Our configuration is adapted from the LIMO example\footnote{\href{https://github.com/GAIR-NLP/LIMO/blob/main/train/examples/train_limo.yaml}{\url{https://github.com/GAIR-NLP/LIMO/blob/main/train/examples/train_limo.yaml}}}. For experiments that augment the LIMO dataset, the \texttt{dataset} field in the configuration should be set to \texttt{limo,sand\_math} assuming the LIMO data\footnote{\href{https://github.com/GAIR-NLP/LIMO/blob/main/train/data/limo.json}{\url{https://github.com/GAIR-NLP/LIMO/blob/main/train/data/limo.json}}} is also available.
    \end{itemize}
    \item \textbf{Launch Training:}
    Execute the training run using the prepared configuration file.
\begin{lstlisting}[language=bash]
llamafactory-cli train \
    examples/train_full/train_sand_math.yaml
\end{lstlisting}

\end{enumerate}

\section{Dataset Characteristics: Consistency, Novelty, and Difficulty}  
\label{app:dataset_characteristics}  

Our multi-stage filtering pipeline produced a high-quality dataset characterized by four key attributes. First, the \textbf{novelty} of the questions was ensured by maintaining an exceptionally low contamination rate of just 0.2\%, measured as the proportion of questions with a similarity score exceeding $\tau = 0.85$ when compared to web search results and existing test sets. Second, a \textbf{self-consistency filtering} stage was applied, requiring that two independently sampled solutions yield the same final answer; this criterion was satisfied by 85\% of the generated question-solution pairs. Third, a \textbf{difficulty filtering} step---dependent on performance by the target model (\texttt{Qwen2.5-32B-Instruct})---was used to retain only 41\% of the consistent questions that were deemed sufficiently challenging. Finally, the \textbf{overall retention rate} across all stages of filtering resulted in a final dataset, $\mathcal{D}_{\text{final}}$, comprising approximately 35\% of the initially generated questions. The difficulty distribution of our dataset SAND-Math was assessed using model-agnostic method (described in Section~\ref{subsec:difficulty_filtering}).

\section{Algorithm for Novelty Filtering}
\label{app:novelty_algorithm}
  
\begin{algorithm}[H] 
\caption{Novelty Filtering Algorithm}  
\label{alg:novelty_check}  
\begin{algorithmic}[1] 
\Require Input dataset $\mathcal{D}_{diff} = \{q_1, q_2, ..., q_N\}$; Similarity threshold $\tau=0.85$; Number of search results $K = 20$; Sentence Transformer model $M_{ST}$; Meta Search Engine $Search(\cdot)$; Final dataset $\mathcal{D}_{final}$

\State $\mathcal{D}_{final} \gets \emptyset$ \Comment{Initialize the final dataset}  
\For{each question $q_i \in \mathcal{D}_{diff}$}  
    \State $Results \gets Search(q_i, K)$ \Comment{Get top K search results (URL, snippet)}  
    \State $max\_similarity \gets 0$  
    \State $q_{emb} \gets M_{ST}.encode(q_i)$ \Comment{Get embedding for the question}  
    \For{each snippet $s_j$ in $Results$}  
        \State $s_{emb} \gets M_{ST}.encode(s_j)$ \Comment{Get embedding for the snippet}  
        \State $similarity \gets cosine\_similarity(q_{emb}, s_{emb})$  
        \If{$similarity > max\_similarity$}  
            \State $max\_similarity \gets similarity$  
        \EndIf  
    \EndFor  
    \If{$max\_similarity \le \tau$} \Comment{Check if highest similarity is below threshold}  
        \State $\mathcal{D}_{final} \gets \mathcal{D}_{final} \cup \{q_i\}$ \Comment{Retain the question}  
    \EndIf  
\EndFor  
\State \Return $\mathcal{D}_{final}$  
\end{algorithmic}  
\end{algorithm}  

\section{Difficulty Hiking Prompt}
\label{app:difficulty_hiking_prompt} 

\begin{tcolorbox}[enhanced jigsaw,breakable,colback=red!5!white,colframe=red!75!black, fonttitle=\bfseries\ttfamily,fontupper=\ttfamily,boxrule=1pt,width=\columnwidth,title={Question difficulty hiking prompt.}]
You are an expert math problem crafter specializing in very hard Olympiad level questions. Your task is to transform the original problem provided below into a new problem with a target difficulty of **challenging Olympiad problem** `(IMO Shortlist level)`. \\

Central Theorem: \{\}  \\
Supporting Concept/Tool: \{\} \\

**Transformation Instructions for the New Problem:** \\

1.  **Deep Synthesis of Concepts:** The solution to the new problem must *critically depend* on the interplay between the original problem's core theme and the newly introduced **Central Theorem**, **Supporting Concept/Tool**. This synthesis should feel natural and integral to the problem. \\

2.  **Reliance on Olympiad-Level Theorem:** The application of **Central Theorem** must be non-trivial, essential for reaching the solution, and demonstrate a deep understanding of the theorem. A superficial application or alternative simpler methods should not suffice. \\

3. **Central Theorem must be disguised:** Central Theorem must be cleaverly disguised. Do not use the **Central Theorem** name in the problem. \\

4.  **Multiple Non-Trivial Intermediate Steps/Lemmas:** Design the problem so its solution requires at least 2-3 distinct, non-obvious intermediate steps. These steps should logically connect the initial problem setup, any necessary lemmas, the application of **Central Theorem**, **Supporting Concept/Tool**, and the derivation of the final answer. \\

5.  **High Degree of Abstraction or Generalization (If Appropriate):** If appropriate, replace concrete numbers from the original with parameters, or frame the question more generally to enhance the conceptual challenge. \\

6.  **Clarity and Soundness:** The new problem statement must be clear, unambiguous, and mathematically sound. \\

7.  **Answer Format:** The problem must be constructed such that it has a **single final integer answer.** \\

**Output Format:** \\
Return the text of the new difficult problem enclosed within `<Q>` and `</Q>` tags and new solution enclosed within <S> and </S>. \\

**Original Problem:** \\
(\textit{\{original\_problem\}})
\end{tcolorbox}

\section{Difficulty Hiking Examples}
\label{app:hiking_example}

Challenging problems are a key component for building high-quality SFT datasets that enhance the mathematical reasoning capabilities of LLMs. Our approach systematically increases problem complexity by transforming an existing problem based on several key transformations such as incorporating new advanced theorem and a supporting concept etc.

Although the LLM is prompted to incorporate all elements, it often applies one of several transformation strategies in isolation. These strategies include:
\begin{itemize}
    \item Incorporating the central theorem.
    \item Incorporating the supporting concept.
    \item Increasing the level of abstraction or generalization.
    \item Adding new constraints.
\end{itemize}

A complete list of the transformation instructions provided to the LLM can be found in the prompt in Appendix~\ref{app:difficulty_hiking_prompt}. This section provides example scenarios for each of the strategies listed above, demonstrating how the LLM applied that transformation to make a problem more difficult.

\begin{tcolorbox}[
    colback=green!5!white,      
    colframe=green!50!black,    
    fonttitle=\bfseries,
    title=Example: Difficulty Hiking using a Central Theorem,
    breakable, 
    pad at break=2mm
]
\textbf{Original Question}

\vspace{0.5em}
\textbf{Question:} Determine the number of distinct real solutions $\theta \in [0, 2\pi)$ satisfying the equations:
\begin{align*}
    \tan(2\theta + \sin\theta) &= \tan(\theta - \sin\theta) \\
    \sin^2\theta + \sin\theta &= 0
\end{align*}

\vspace{0.5em}
\textbf{Branch:} Algebra \\
\textbf{Initial Difficulty:} 2.0 / 10.0

\tcbline 

\textbf{Transformation Inputs}

\vspace{0.5em}
\textbf{Theorem (from same branch):} Cauchy's Functional Equation ($f(x+y)=f(x)+f(y)$) (and its solutions over $\mathbb{Q}$, $\mathbb{R}$ with conditions) \\
\textbf{Concept:} Sprague-Grundy Theorem (Nim-values, for impartial games -- conceptual awareness) \\
\textbf{Target Difficulty:} 8.0

\tcbline 

\textbf{Hiked Question (Output)}
\vspace{0.5em}

\textbf{Question:} Consider a function $f: \mathbb{R} \to \mathbb{R}$ that satisfies the following conditions:
\begin{enumerate}
    \item $f(x + y) = f(x) + f(y)$ for all real numbers $x$ and $y$.
    \item $f(\pi) = 0$.
    \item For all $\theta \in [0, 2\pi)$, the equation $f(2\theta) = 2f(\theta) + f(\sin^2 \theta)$ holds.
\end{enumerate}

Additionally, consider an impartial combinatorial game played on the set of non-negative integers. The rules are:
\begin{itemize}
    \item From a position $n \geq 0$, a player can move to any integer $m$ such that $0 \leq m < n$ and $f(m) = f(n) - 1$.
    \item The position $0$ is terminal (no moves).
\end{itemize}

Let $\omega(k)$ denote the Grundy number (nimber) of position $k$ in this game. It is known that $\omega$ is eventually periodic.

Determine the number of integers $k$ in the set $\{0, 1, 2, \dots, 2024\}$ for which $\omega(k) = 0$.
\vspace{0.5em}

\textbf{New Difficulty:} 7.0 / 10.0

\tcbline 

\textbf{Answer Analysis}

\vspace{0.5em}
The problem can be solved by first determining the specific function $f$ that satisfies the given conditions, and then analyzing the combinatorial game based on this function. The specified theorem, \textbf{Cauchy's Functional Equation}, is the cornerstone for finding the function $f$. . . . . . . . . . . . . Since this is true for all $k$, we just need to count the number of elements in the set. The set contains $2024 - 0 + 1 = 2025$ integers.

\end{tcolorbox}


\begin{tcolorbox}[
    colback=green!5!white,      
    colframe=green!50!black,    
    fonttitle=\bfseries,
    title=Example: Difficulty Hiking using a Supporting Concept,
    breakable, 
    pad at break=2mm
]
\textbf{Original Question}

\vspace{0.5em}
\textbf{Question:} Let $p$ be an odd prime. Find the number of subsets $S \subseteq \{1, 2, \ldots, p-1\}$ such that both the sum and the product of the elements of $S$ are congruent to $0$ modulo $p$.

\vspace{0.5em}
\textbf{Branch:} Number Theory \\
\textbf{Initial Difficulty:} 1.0 / 10.0

\tcbline 

\textbf{Transformation Inputs}

\vspace{0.5em}
\textbf{Theorem (from same branch):} Euler's Totient Theorem \\
\textbf{Concept:} Cyclic Groups (Conceptual understanding, e.g., $(\mathbb{Z}/p\mathbb{Z})^*$ is cyclic)\\
\textbf{Target Difficulty:} 8.0

\tcbline 

\textbf{Hiked Question (Output)}
\vspace{0.5em}

\textbf{Question:} Let $n = 9$. Consider the set $G = \{ k \mid 1 \leq k \leq 8, \gcd(k, 9) = 1 \}$. Find the number of subsets $S \subseteq G$ such that the sum of the elements in $S$ is divisible by 9, and the product of the elements in $S$ is congruent to 1 modulo 9.

\vspace{0.5em}
\textbf{New Difficulty:} 6.0 / 10.0

\tcbline 

\textbf{Answer Analysis}

\vspace{0.5em}
The problem asks for the number of subsets $S$ of the set $G = \{ k \mid 1 \leq k \leq 8, \gcd(k, 9) = 1 \}$ that satisfy two conditions: the sum of elements is divisible by 9, and the product of elements is congruent to 1 modulo 9.

\subsubsection*{Step 1: Identify G using the concept of Cyclic Groups}

First, we identify the elements of the set $G$ and analyze its structure as a cyclic group under multiplication modulo 9. . . . . . . . . . . . . . . 
The total number of such subsets is 2.

The use of the \textbf{cyclic group concept} was essential to transform the difficult product condition $\prod_{s \in S} s \equiv 1 \pmod{9}$ into a much simpler linear congruence on the exponents, $\sum i \equiv 0 \pmod{6}$, making the problem systematic and solvable.
\end{tcolorbox}


\begin{tcolorbox}[
    colback=green!5!white,      
    colframe=green!50!black,    
    fonttitle=\bfseries,
    title=Example: Difficulty Hiking by increasing Abstraction,
    breakable, 
    pad at break=2mm
]
\textbf{Original Question}

\vspace{0.5em}
\textbf{Question:} A bag contains red and blue balls. When two balls are drawn with replacement, the probability that both are red is $\frac{1}{4}$. Additionally, the probability of drawing at least one red ball in two draws is $\frac{3}{4}$. Determine the total number of balls in the bag.

\vspace{0.5em}
\textbf{Branch:} Probability \\
\textbf{Initial Difficulty:} 2.0 / 10.0

\tcbline 

\textbf{Transformation Inputs}

\vspace{0.5em}
\textbf{Theorem (from same branch):} Lovasz Local Lemma (Tool for showing existence when events are \"mostly\" independent - very advanced). \\
\textbf{Concept:} Infinite Descent Method (Fermat's Method). \\
\textbf{Target Difficulty:} 8.0

\tcbline 

\textbf{Hiked Question (Output)}
\vspace{0.5em}

\textbf{Question:} Determine the smallest positive integer $n$ such that for any assignment of the color red or blue to each integer in the set $\{1, 2, \dots, n\}$, there is an arithmetic progression of length 4 that is monochromatic (all red or all blue)

\vspace{0.5em}
\textbf{New Difficulty:} 7.0 / 10.0

\tcbline 

\textbf{Higher Abstraction Explanation}

\vspace{0.5em}
The original problem is a concrete probability calculation involving specific numerical probabilities (1/4 and 3/4) in a single scenario (one bag of balls). The new problem generalizes this by asking for a universal property that must hold for *any* assignment of colors (a class of functions) on a set of integers {1, 2, ..., n}. It replaces concrete calculations with a search for a threshold parameter 'n' where a specific combinatorial structure (a monochromatic arithmetic progression) is guaranteed to exist, which represents a significant increase in abstraction and conceptual challenge.
\end{tcolorbox}

\begin{tcolorbox}[
    colback=green!5!white,      
    colframe=green!50!black,    
    fonttitle=\bfseries,
    title=Example: Difficulty Hiking using a Central Theorem,
    breakable, 
    pad at break=2mm
]
\textbf{Original Question}
\vspace{0.5em} 

\textbf{Question:} Let \(n\) be a positive integer. Determine the number of positive integers \(n\) for which
\[
    \frac{\sec^2(\pi n)}{\tan^2(\pi n)} + \frac{\sin(\pi n)}{2}
\]
is a non-negative integer.

\vspace{0.5em}
\textbf{Branch:} Number Theory \\
\textbf{Initial Difficulty:} 1.0 / 10.0

\tcbline 

\textbf{Transformation Inputs}
\vspace{0.5em}

\textbf{Theorem (from same branch):} M\"obius Inversion Formula \\
\textbf{Concept:} Primitive Roots (Existence criteria, properties) \\
\textbf{Target Difficulty:} 8.0

\tcbline 

\textbf{Hiked Question (Output)}
\vspace{0.5em}

\textbf{Question:} Let \(n\) be a square-free positive integer for which there exists a primitive root modulo \(n\). Determine the number of such \(n\) for which the sum
\[
    \sum_{\substack{k=1 \\ \gcd(k,n)=1}}^n \sin^2\left(\frac{\pi k}{n}\right)
\]
is an integer.

\vspace{0.5em}
\textbf{New Difficulty:} 7.0 / 10.0
\end{tcolorbox}

\section{Question Generation Prompt}
\label{app:question_prompt} 

In this section, we present the prompt used for question generation, which is shown below. \texttt{\{primary \_math\_branch\}} and \texttt{\{secondary\_math\_branch\}} are randomly selected from the list of mathematics branches that generally appear in Olympiad-level competitions.

\begin{tcolorbox}[enhanced jigsaw,breakable,colback=blue!5!white,colframe=blue!75!black, fonttitle=\bfseries\ttfamily,fontupper=\ttfamily,boxrule=1pt,width=\columnwidth,title={Question Generation Prompt}]
Generate one novel math problem and solution with a difficulty level at National or International Olympiads.\\

It must have a single non-negative integer as the answer.  \\

The problem should primarily focus on (\textit{\{primary\_math\_branch\}}) and incorporate a clever mix of elements from (\textit{\{secondary\_math\_branch\}}).\\       
  
Your response should be formatted as follows: \\ 
<Q> Problem Statement </Q>  \\
<S> Step-by-step solution, concluding with the final answer enclosed in boxed\{\}. </S>  \\
        
\end{tcolorbox}

\section{Solution Generation Prompt}
\label{app:solution_prompt} 
\begin{tcolorbox}[enhanced jigsaw,breakable,colback=purple!5!white,colframe=purple!75!black, fonttitle=\bfseries\ttfamily,fontupper=\ttfamily,boxrule=1pt,width=\columnwidth,title={Question Generation Prompt}]
Answer the mathematics question. Think step by step and put your final answer with in boxed\{\} \\

question:\\
\{Math question\} \\
        
\end{tcolorbox}

\section{Question Difficulty Rating Prompt}
\label{app:diffclass_prompt} 

\begin{tcolorbox}[enhanced jigsaw,breakable,colback=purple!5!white,colframe=purple!75!black, fonttitle=\bfseries\ttfamily,fontupper=\ttfamily,boxrule=1pt,width=\columnwidth,title={Question difficulty rating prompt.}]
\# ROLE: Expert Math Problem Difficulty Assessor \\

\# TASK: \\
Analyze the provided math problem and solution to assign a difficulty score based on the provided reference materials. \\

\# REFERENCE MATERIALS: \\
<difficulty\_reference> \\
\#\# Difficulty Level Descriptions (1.0 - 10.0 Scale)  \\ 

    1.0: Problems strictly for beginner, on the easiest elementary school or middle school levels (MOEMS, MATHCOUNTS School, AMC 8 1-10, AMC 10 1-10, easier AMC 12 1-5, and others that involve standard techniques introduced up to the middle school level), most traditional middle/high school word problems.   \\ 

    1.5: Problems for stronger beginner students, on the level of the middling problems in most middle school contests (AMC 8 11-20, harder AMC 10 1-10, AMC 12 1-5, and those others that force students to apply their school-level knowledge to slightly more challenging problems), traditional middle/high school word problems with more complex problem solving.  \\ 

    2.0: For motivated beginners, harder questions from the previous categories (AMC 8 21-25, MATHCOUNTS Chapter (Sprint 21-30, Target 6-8), MATHCOUNTS States/Nationals, AMC 10 11-15, AMC 12 5-10, easiest AIME 1-3)  \\ 

    2.5: More advanced beginner problems, hardest questions from previous categories (Harder AMC 8 21-25, harder MATHCOUNTS States questions, AMC 10 16-20, AMC 12 11-15, usual AIME 1-3)  \\ 

    3.0: Early intermediate problems that require more creative thinking (harder MATHCOUNTS National questions, AMC 10 21-25, AMC 12 15-20, hardest AIME 1-3, usual AIME 4-6).  \\ 

    3.5: Problems requiring non-trivial insights or synthesis (Harder AMC 10 21-25, harder AMC 12 15-20, usual AIME 4-6, harder AIME 7-9).  \\ 

    4.0: Intermediate-level problems (AMC 12 21-25, hardest AIME 4-6, usual AIME 7-10).  \\ 

    4.5: Strong intermediate problems bridging to Olympiad style (Harder AMC 12 21-25, usual AIME 10-12, easiest USAJMO 1/4).  \\ 

    5.0: More difficult AIME problems (11-13), simple proof-based Olympiad-style problems (early JBMO questions, easier USAJMO 1/4).  \\ 

    5.5: Challenging AIME problems, introductory Olympiad proofs (Hardest AIME 11-13, average USAJMO 1/4, easier USAJMO 2/5).  \\ 

    6.0: High-leveled AIME-styled questions (14/15). Introductory-leveled Olympiad-level questions (harder USAJMO 1/4 and easier USAJMO 2/5, easier USAMO and IMO 1/4).  \\ 

    6.5: Solid introductory Olympiad problems (Average USAJMO 2/5, harder USAJMO 3/6, average USAMO/IMO 1/4).  \\ 

    7.0: Tougher Olympiad-level questions, may require more technical knowledge (harder USAJMO 2/5 and most USAJMO 3/6, extremely hard USAMO and IMO 1/4, easy-medium USAMO and IMO 2/5).  \\ 

    7.5: Strong Olympiad problems (Hardest USAJMO 3/6, medium USAMO/IMO 2/5).  \\ 

    8.0: High-level Olympiad-level questions (medium-hard USAMO and IMO 2/5, easiest USAMO and IMO 3/6).  \\ 

    8.5: Very challenging Olympiad problems (Hard USAMO/IMO 2/5, average USAMO/IMO 3/6).  \\ 

    9.0: Expert Olympiad-level questions (average USAMO and IMO 3/6).  \\ 

    9.5: The hardest problems appearing on Olympiads which the strongest students could reasonably solve (hard USAMO and IMO 3/6).  \\ 

    10.0: Historically hard problems, generally unsuitable for very hard competitions (such as the IMO) due to being exceedingly tedious, long, and difficult (e.g. very few students are capable of solving on a worldwide basis).  \\ 
   
    </difficulty\_reference>  \\

\# INSTRUCTIONS: \\
1.  **Analyze:** Carefully read the provided `Math Problem` and its `Solution`. Identify the core mathematical concepts, required techniques, and the complexity of the argument. Note any particularly clever steps, non-obvious insights, or reliance on advanced theorems. \\

2.  **Compare:** Compare the analyzed problem to the `Difficulty Level Descriptions` provided in the reference materials. Consider where it fits in terms of typical competition level (AMC 8/10/12, AIME, USA(J)MO, IMO) and the type of thinking required. \\

3.  **Score:** Assign a difficulty score between **1.0 and 10.0**, using increments of **0.5** (e.g., 3.0, 3.5, 4.0). The score must be consistent with the provided reference scale. \\

4.  **Summarize:** Write a brief paragraph summarizing the problem's core topic and mathematical area(s). Enclose this summary within `<S>` and `</S>` tags. \\

5.  **Assign Score:** Place the difficulty score assigned in step 3 within `<D>` and `</D>` tags. \\

6.  **Justify:** Write a paragraph explaining the reasoning behind the assigned difficulty score, explicitly referencing the comparison made in step 2 (e.g., "This problem involves techniques similar to example 3.5..." or "The required insight aligns with the description for level 6.0..."). Mention aspects of the problem or solution (like multi-step reasoning, specific theorems, type of creativity needed) that justify the score. Enclose this justification within `<R>` and `</R>` tags. \\

\# OUTPUT FORMAT: \\
<S>[Your brief paragraph summarizing the problem.]</S> \\
<D>[The assigned score, e.g., 4.5]</D> \\
<R>[Your paragraph justifying the score based on analysis and comparison to references.]</R> \\

\# INPUT PROBLEM \& SOLUTION: \\

\#\# Math Problem: \\
<|question|> \\

\#\# Solution: \\
<|solution|>
        
\end{tcolorbox}

\section{Question Decontamination Prompt}
\label{app:decontamination_prompt} 

\begin{tcolorbox}[enhanced jigsaw,breakable,colback=red!5!white,colframe=red!75!black, fonttitle=\bfseries\ttfamily,fontupper=\ttfamily,boxrule=1pt,width=\columnwidth,title={Question decontaminaiton prompt.}]
Determine whether the provided new question is identical to or a paraphrased version of any of the existing questions listed. 
If it is identical or paraphrased, respond with **yes** otherwise, respond with **no**.  \\
Please ensure your response is only yes or no, with no additional commentary. \\

New Question: \\
(\textit{\{synthetic\_question\}}) \\

Existing Questions: \\
(\textit{\{list\_of\_similiar\_questions\}})
\end{tcolorbox}

\section{Math Theorems and Concepts Taxonomy}
\label{app:math_tax}

\begin{tcolorbox}[enhanced jigsaw,breakable,colback=black!5!white,colframe=black!75!black, fonttitle=\bfseries\ttfamily,fontupper=\ttfamily,boxrule=1pt,width=\columnwidth,title={Taxonomy}]
Number Theory: \\
  \# topic \\
  Divisibility and Prime Factorization: \\
    tools\_concepts: \\ 
      - Division Algorithm ($a = bq + r$) \\
      - Greatest Common Divisor (GCD) \& Least Common Multiple (LCM) \\
        (Properties, gcd*lcm = |ab|) \\
      - Euclidean Algorithm (For GCD and linear Diophantine solutions) \\
      - Bézout's Identity ($ax + by = \text{gcd}(a,b)$) \\
      - Prime Numbers \& Composite Numbers (Definitions, Sieve of Eratosthenes) \\
      - Fundamental Theorem of Arithmetic (Unique Prime Factorization) \\
      - p-adic Valuation ($v_p(n)$) \\
      - Legendre's Formula (for $v_p(n!)$) \\
    theorems: \\
      - Euclid's Theorem on Infinitude of Primes \\
      - Dirichlet's Theorem on Arithmetic Progressions (Existence Statement) \\
 
  Modular Arithmetic: \\
    tools\_concepts: \\
      - Congruence Relation ($\equiv$) (Properties) \\
      - Complete Residue System (CRS)\& Reduced Residue System (RRS) \\
      - Linear Congruences ($ax \equiv b \pmod m$) 
      (Solvability, number of solutions) \\
      - Modular Inverse (Existence and calculation) \\
      - Order of an Element modulo n ($\text{ord}_n(a)$)  (Properties) \\
      - Primitive Roots (Existence criteria, properties) \\
      - Quadratic Residues\& Non-Residues (Definition) \\
      - Legendre Symbol ($(\frac{a}{p})$) (Definition and properties) \\
    Theorems: \\
      - Fermat's Little Theorem (FLT) \\
      - Euler's Totient Theorem \\
      - Wilson's Theorem \\
      - Chinese Remainder Theorem (CRT) (Solvability and construction) \\
      - Lagrange's Theorem (for polynomial roots modulo p) \\
      - Lifting The Exponent Lemma (LTE) \\
      - Euler's Criterion \\
      - Law of Quadratic Reciprocity  \\
        (and properties for $(\frac{-1}{p})$, $(\frac{2}{p})$) \\
 
  Diophantine Equations: \\
    tools\_concepts: \\
      - Linear Diophantine Equations ($ax+by=c$) (Structure of solutions) \\
      - Pythagorean Triples ($x^2+y^2=z^2$) (Parametrization) \\
      - Pell's Equation ($x^2 - Dy^2 = 1$)
        (Structure of solutions, fundamental solution) \\
      - Factoring Techniques for Diophantine Equations 
        (Difference of squares, sum/difference of cubes, etc.) \\
      - Modular Arithmetic Constraints for Diophantine Equations 
        (Proving no solutions) \\
      - Infinite Descent Method (Fermat's Method) \\
      - Vieta Jumping Technique \\
      - Bounding/Ordering Variables in Diophantine Equations
    theorems: \\
      - Thue's Theorem (Finiteness of solutions - conceptual awareness) \\
      - Catalan's Conjecture (Mihailescu's Theorem) 
        (Specific unique solution $3^2 - 2^3 = 1$) \\
 
  Number Theoretic Functions: \\
    tools\_concepts: \\
      - $\phi(n)$ (Euler's Totient Function) (Formula, multiplicativity) \\
      - $d(n)$ or $\tau(n)$ (Number of Divisors Function) 
        (Formula, multiplicativity) \\
      - $\sigma(n)$ (Sum of Divisors Function) 
        (Formula, multiplicativity, $\sigma_k(n)$) \\
      - $\mu(n)$ (Möbius Function) (Definition, multiplicativity) \\
      - Floor Function ($\lfloor x \rfloor$)\& Fractional Part ($\{x\}$) \\
        (Properties)
      - Definitions of Perfect Numbers, Amicable Numbers
    theorems: \\
      - Möbius Inversion Formula \\
 
  Polynomials in Number Theory: \\
    tools\_concepts: \\
      - Integer-valued polynomials (Properties) \\
      - Cyclotomic Polynomials ($\Phi_n(x)$) (Definition, properties, values) \\
    theorems: \\
      - Rational Root Theorem \\
      - Gauss's Lemma (on polynomial content and irreducibility) \\
      - Eisenstein's Criterion (for irreducibility over $\mathbb{Q}$) \\
 
Algebra: \\
  Polynomials: \\ 
    tools\_concepts: \\
      - Polynomial Long Division and Remainder Theorem \\
      - Factor Theorem and Root Theorem \\
      - Vieta's Formulas (Relating roots and coefficients) \\
      - Symmetric Sums of Roots (Expressing symmetric polynomials 
        in terms of elementary symmetric polynomials) \\
      - Properties of Polynomial Roots (Real, complex, conjugate pairs) \\
      - Divisibility of Polynomials \\
      - Polynomial Interpolation (e.g., Lagrange Interpolation concept) \\
      - Integer Roots and Rational Root Theorem \\
      - Content of a Polynomial \\
      - Cyclotomic Polynomials (Algebraic properties, connection to roots of unity) \\
    theorems: \\
      - Fundamental Theorem of Algebra (Existence of roots in $\mathbb{C}$) \\
      - Newton's Sums (Relating power sums of roots and elementary 
        symmetric polynomials) \\
      - Gauss's Lemma (on polynomial content and irreducibility over $\mathbb{Q}$) \\
      - Eisenstein's Criterion (for irreducibility over $\mathbb{Q}$) \\
      - Lucas's Theorem (for binomial coefficients modulo p - often used with polynomials over finite fields)
 
  Inequalities: \\
    tools\_concepts: \\
      - Basic Inequality Properties (Transitivity, addition/multiplication by constants) \\
      - Completing the Square \\
      - Trivial Inequality ($x^2 \ge 0$) \\
      - Rearrangement of Terms / Substitution Techniques \\
      - Homogenization and Normalization \\
      - Convexity/Concavity of Functions (Conceptual basis for Jensen's) \\
      - Smoothing Principle (Reducing variables or making terms closer) \\
    theorems: \\ 
      - AM-GM Inequality (Arithmetic Mean - Geometric Mean)\\
      - GM-HM Inequality (Geometric Mean - Harmonic Mean)\\
      - Weighted AM-GM Inequality\\
      - Cauchy-Schwarz Inequality (Engel form, Titu's Lemma)\\
      - Rearrangement Inequality\\
      - Jensen's Inequality (for convex/concave functions)\\
      - Muirhead's Inequality (for comparing symmetric sums)\\
      - Schur's Inequality\\
      - Holder's Inequality\\
      - Minkowski's Inequality\\
      - Nesbitt's Inequality (Specific common Olympiad inequality)\\
 
  Functional Equations:\\
    tools\_concepts:\\
      - Substitution of Specific Values (e.g., x=0, y=1, y=x, y=-x)\\
      - Checking for Injectivity, Surjectivity, Bijectivity\\
      - Finding Fixed Points ($f(x)=x$)\\
      - Exploiting Symmetry\\
      - Iteration of the function ($f(f(x))$, etc.)\\
      - Reduction to Known Equations (e.g., Cauchy forms)\\
      - Assuming properties (continuity, differentiability) \\
        to find candidate solutions (then verifying for all reals if needed)\\
      - Domain and Range Analysis\\
    theorems: \# Often, the "theorems" are the well-known solutions to standard forms\\
      - Cauchy's Functional Equation ($f(x+y)=f(x)+f(y)$) 
        (and its solutions over $\mathbb{Q}$, $\mathbb{R}$ with conditions)\\
      - Jensen's Functional Equation ($f(\frac{x+y}{2}) = \frac{f(x)+f(y)}{2}$)\\
      - D'Alembert's Functional Equation (Cosine form: $f(x+y)+f(x-y)=2f(x)f(y)$)\\
      - D'Alembert's Functional Equation (Sine form: $f(x+y)f(x-y)=f(x)^2-f(y)^2$)\\
 
  Sequences and Series:\\
    tools\_concepts:\\
      - Arithmetic Progressions (AP) (Definition, sum formula)\\
      - Geometric Progressions (GP) (Definition, sum formula, infinite GP sum)\\
      - Recurrence Relations (Definition, finding terms)\\
      - Linear Homogeneous Recurrence Relations with Constant Coefficients 
        (Method of characteristic equation)\\
      - Linear Non-Homogeneous Recurrence Relations\\
      - Telescoping Sums and Products\\
      - Bounding Sequences (Monotonicity, boundedness)\\
      - Summation Techniques ($\sum k, \sum k^2, \sum k^3$)\\
      - Difference Operator / Finite Calculus (less common, but a tool)\\
    theorems:\\
      - Binet's Formula (for Fibonacci numbers - example of solving a recurrence)\\
      - Master Theorem for divide-and-conquer recurrences 
        (more CS, but spirit can appear)\\
     \# Specific convergence/divergence tests are usually beyond Olympiad scope
        unless very elementary\\
 
  Abstract Algebra (Elements relevant to Olympiads):\\
    tools\_concepts: \# Focus on concrete applications rather than deep theory\\
      - Basic Group Properties (Closure, associativity, identity, inverse - often
        in modular arithmetic or transformations)\\
      - Cyclic Groups (Conceptual understanding, e.g., 
        $(\mathbb{Z}/p\mathbb{Z})^*$ is cyclic)\\
      - Lagrange's Theorem (Order of subgroup divides order of group \- often 
        used in Number Theory contexts)\\
      - Group Actions on Sets 
        (Conceptual basis for Burnside's Lemma/Polya Enumeration)\\
      - Basic Ring and Field Properties (e.g., $\mathbb{Z}_p$ is a field,
        properties of polynomial rings - conceptual)\\
    theorems:\\
      - Cayley's Theorem 
        (Every group is isomorphic to a group of permutations - conceptual)\\
     \# Most deep theorems from abstract algebra are beyond typical Olympiad scope, 
        but their spirit or basic consequences can be used.\\
 
Combinatorics:\\
  Basic Counting Principles:\\
    tools\_concepts:\\
      - Addition Principle (Rule of Sum)\\
      - Multiplication Principle (Rule of Product)\\
      - Permutations ($P(n,k)$)\\
      - Combinations ($\binom{n}{k}$)
        (Properties of binomial coefficients, Pascal's Identity)\\
      - Stars and Bars (Combinations with repetition, solutions to $x_1+...+x_k=n$)\\
      - Casework and Complementary Counting (Strategy)\\
      - Bijection Principle (Counting one set by mapping to another)\\
      - Double Counting (Establishing an identity by counting a quantity in two ways)\\
    theorems:\\
      - Binomial Theorem ($(x+y)^n = \sum \binom{n}{k}x^k y^{n-k}$)\\
      - Multinomial Theorem\\
      - Vandermonde's Identity (for binomial coefficients)\\
      - Lucas's Theorem (for $\binom{n}{k} \pmod p$)\\
 
  Advanced Counting Techniques:\\
    tools\_concepts:\\
      - Principle of Inclusion-Exclusion (PIE) (General formula and applications)\\
      - Derangements (!n or $D_n$)\\
      - Recursion and Recurrence Relations (Setting up combinatorial recurrences)\\
      - Generating Functions (Ordinary - OGF) \\
        (Representing sequences, solving recurrences, coefficient extraction)\\
      - Exponential Generating Functions (EGF) (For labeled objects, permutations)\\
      - Rook Polynomials \\
        (For counting placements on a board with restrictions - less common)\\
    theorems:\\
     \# Many "theorems" here are specific results derived using these techniques
        or identities related to generating functions.\\
 
  Graph Theory:\\
    tools\_concepts:\\
      - Basic Definitions 
        (Vertices, edges, degree, directed/undirected graphs, weighted graphs)\\
      - Paths, Cycles, Connectedness, Components\\
      - Trees (Properties: $n-1$ edges, unique paths, acyclic and connected)\\
      - Bipartite Graphs (Characterization: no odd cycles)\\
      - Adjacency Matrix and Incidence Matrix (Representations)\\
      - Graph Isomorphism\\
      - Eulerian Paths and Circuits (Conditions based on vertex degrees)\\
      - Hamiltonian Paths and Cycles 
        (No simple general condition, often NP-complete - problems focus on specific 
        cases or properties)\\
      - Planar Graphs (Definition, Kuratowski's Theorem - conceptual awareness)\\
      - Graph Coloring (Vertex coloring, chromatic number, edge coloring)\\
      - Matchings in Graphs (e.g., Hall's Marriage Theorem condition)\\
      - Network Flows (Max-flow min-cut theorem - conceptual)\\
    theorems:\\
      - Handshaking Lemma ($\sum \text{deg}(v) = 2|E|$)\\
      - Euler's Formula for Planar Graphs ($V-E+F=2$)\\
      - Mantel's Theorem (Max edges in a triangle-free graph)\\
      - Turan's Theorem (Generalization of Mantel's Theorem - extremal graph theory)\\
      - Hall's Marriage Theorem (Condition for perfect matching in bipartite graphs)\\
      - Cayley's Formula \\
        (Number of spanning trees in $K_n$ is $n^{n-2}$)\\
      - Ramsey's Theorem 
        (Existence of monochromatic cliques in edge-colored complete graphs, 
        e.g., $R(3,3)=6$)\\
 
  Combinatorial Designs\& Extremal Combinatorics:\\
    tools\_concepts:\\
      - Set Systems (Families of subsets)\\
      - Intersecting Families, Antichains\\
      - Pigeonhole Principle (PHP)
      (Simple and generalized forms, advanced applications)\\
      - Extremal Principle \\
      (Considering objects with maximal/minimal properties)\\
      - Design Theory Basics \\
      (e.g., Balanced Incomplete Block Designs - BIBD - conceptual)\\
      - Latin Squares, Orthogonal Latin Squares\\
    theorems:\\
      - Sperner's Theorem (Max size of an antichain)\\
      - Dilworth's Theorem (Relating antichains and chain decompositions in posets)\\
      - Erdos-Ko-Rado Theorem (Max size of an intersecting family of k-subsets)\\
     \# Many results in this area are specific extremal bounds rather than general
        named theorems taught widely.\\
 
  Probabilistic Method\& Combinatorial Probability:\\
    tools\_concepts:\\
      - Basic Probability 
        (Sample space, events, independence, conditional probability)
      - Expected Value (Linearity of Expectation)\\
      - Indicator Variables 
        (Random variables, powerful tool with linearity of expectation)\\
      - Basic Probabilistic Method (Showing existence by proving probability > 0)\\
      - Alteration Method (Modifying a random structure to get desired properties)\\
      - Lovasz Local Lemma (Advanced tool for rare events - conceptual awareness)\\
    theorems:\\
     \# Results are often existence proofs rather than named theorems, e.g.,
        "There exists a graph with property X..."
     \# Markov's Inequality, Chebyshev's Inequality (Basic probabilistic bounds)\\
 
  Combinatorial Games\& Processes:\\
    tools\_concepts:\\
      - Invariants (Quantities that remain unchanged during a process)\\
      - Monovariants (Quantities that strictly increase or decrease)\\
      - Winning and Losing Positions (P-positions, N-positions)\\
      - Symmetry Arguments in Games\\
      - Strategy Stealing Arguments\\
      - Sprague-Grundy Theorem 
        (Nim-values, for impartial games - conceptual awareness)
      - Coloring Arguments (e.g., tiling problems)\\
    theorems:\\
     \# Often specific to the game or process being analyzed.\\
     \# Zermelo's Theorem (Existence of a winning strategy in finite, 
        perfect information, two-player games with no draws)\\
 
Geometry:\\
  Triangle Geometry (Advanced):\\
    tools\_concepts:\\
      - Special Points (Centroid, Incenter, Circumcenter, Orthocenter, Excenters,
        Nagel Point, Gergonne Point, Isodynamic Points, Isogonal Conjugates, 
        Symmedian Point/Lemoine Point) and their properties/collinearities/concurrencies.\\
      - Euler Line and Nine-Point Circle (Properties and relations)\\
      - Simson Line (and its generalizations)\\
      - Pedal Triangles\\
      - Cevas Theorem (Trigonometric, standard, and converse forms)\\
      - Menelaus Theorem (Standard and converse forms)\\
      - Angle Bisector Theorem (Internal and External)\\
      - Stewart's Theorem\\
      - Routh's Theorem (for ratios in cevians)\\
      - Properties of Medians, Altitudes, Angle Bisectors (lengths, intersections)\\
      - Isogonal Conjugacy (Definition and key properties)\\
      - Isotomic Conjugacy (Definition and key properties)\\
      - Brocard Points and Brocard Angle\\
    theorems:\\
      - Morley's Trisector Theorem\\
      - Napoleon's Theorem\\
      - Van Aubel's Theorem\\
      - Feuerbach's Theorem (Nine-point circle tangency to incircle/excircles)\\
      - Lester's Theorem 
        (Circumcenter, nine-point center, and Brocard points are concyclic)\\
 
  Circles and Cyclic Quadrilaterals:\\
    tools\_concepts:\\
      - Power of a Point Theorem (for intersecting secants, tangents, chords)\\
      - Radical Axis (of two circles, properties, construction)\\
      - Radical Center (of three circles)\\
      - Coaxal Circles (Pencils of circles)\\
      - Properties of Cyclic Quadrilaterals 
        (Opposite angles sum to 180, external angle property)\\
      - Properties of Tangential Quadrilaterals (Opposite sides sum equally)\\
      - Ptolemy's Theorem and Ptolemy's Inequality 
        (for cyclic and general quadrilaterals)\\
      - Casey's Theorem (Generalized Ptolemy's Theorem for tangent circles)\\
      - Directed Angles 
        (modulo $\pi$ or $2\pi$, for handling configurations carefully)\\
    theorems:\\
      - Brahmagupta's Formula (Area of a cyclic quadrilateral)\\
      - Japanese Theorem for Cyclic Quadrilaterals\\
      - Miquel's Theorem and Miquel Point 
        (for complete quadrilaterals or triangles with points on sides)\\
 
  Geometric Transformations:\\
    tools\_concepts:\\
      - Homothety (Dilation) (Properties, center of homothety, composition)\\
      - Rotation (Properties, center of rotation, composition)\\
      - Reflection (Properties, composition creating rotations/translations)\\
      - Translation\\
      - Spiral Similarity (Composition of homothety and rotation)\\
      - Inversion (Properties: maps circles/lines to circles/lines, preserves angles 
      (conformal), changes distances in a specific way, center of inversion)\\
      - Glide Reflection\\
    theorems:\\
     \# Many "theorems" are properties preserved or created by these transformations.\\
     \# e.g., "Homothety maps a line to a parallel line."\\
     \# e.g., "Inversion preserves cross-ratios of four concyclic points if center
        of inversion is also on the circle." (Special case)\\
 
  Analytic and Vectorial Geometry (as Tools):\\
    tools\_concepts:\\
      - Cartesian Coordinates (Distance, slope, midpoint, equation of line/circle)\\
      - Vector Addition, Subtraction, Scalar Multiplication\\
      - Dot Product (for angles, perpendicularity, lengths)\\
      - Cross Product (for area, normal vectors, collinearity in 3D, orientation)\\
      - Barycentric Coordinates (Representing points in a triangle, 
        proving concurrency/collinearity, area ratios)\\
      - Complex Numbers in Geometry (Representing points, vectors, rotations,
        similarities, conditions for collinearity/concyclicity, roots of unity for 
        regular polygons)\\
      - Distance Formulas (Point-line, point-plane)\\
    theorems:\\
     \# Results are often derived using these tools rather than being standalone 
        theorems from this "topic" for Olympiads.\\
     \# e.g., Shoelace Formula (for area of polygon using coordinates).\\
     \# e.g., Conditions for collinearity/concurrency using determinants or vector
        dependencies.\\
 
  Projective Geometry (Elements for Olympiads):\\
    tools\_concepts:\\
      - Points at Infinity, Line at Infinity\\
      - Duality Principle\\
      - Cross-Ratio (of four collinear points or four concurrent lines, invariance
        under projection)\\
      - Harmonic Bundles and Harmonic Conjugates (Cross-ratio = -1, 
        geometric constructions)\\
      - Perspective Triangles\\
      - Complete Quadrilaterals and Complete Quadrangles (Harmonic properties)\\
      - Poles and Polars (with respect to a conic, especially a circle)\\
    theorems:\\
      - Desargues' Theorem (Perspective triangles from a point and a line)\\
      - Pappus's Hexagon Theorem\\
      - Pascal's Theorem (for hexagons inscribed in a conic)\\
      - Brianchon's Theorem (for hexagons circumscribed about a conic)\\
     \# La Hire's Theorem\\
 
  Solid Geometry (Euclidean 3D):\\
    tools\_concepts:\\
      - Lines and Planes in 3D (Intersection, parallelism, perpendicularity, 
        skew lines)\\
      - Dihedral Angles, Solid Angles\\
      - Properties of Basic Solids (Prisms, pyramids, cylinders, cones, spheres)\\
      - Euler's Formula for Polyhedra ($V-E+F=2$)\\
      - Regular Polyhedra (Platonic solids, properties)\\
      - Coordinate Geometry in 3D (Distance, equations of planes/lines, 
        vector operations)\\
      - Cavalieri's Principle (for volumes)\\
    theorems:\\
     \# Fewer named "theorems" at Olympiad level compared to 2D; more about spatial 
      reasoning and application of 2D principles in planes.\\
     \# e.g., Relationship between slant height, height, and base radius of a cone.\\
     \# Desargues' Theorem can be proven using 3D perspective.\\
 
  Combinatorial Geometry:\\
    tools\_concepts:\\
      - Convex Hull (Properties, algorithms like Graham scan - conceptual)\\
      - Geometric Inequalities (e.g., Triangle inequality, Ptolemy's inequality,
        Isoperimetric inequality - basic form)\\
      - Covering, Packing, and Tiling Problems (Basic examples)\\
      - Incidence Problems (e.g., Sylvester-Gallai problem spirit)\\
      - Coloring Geometric Configurations\\
      - Helly's Theorem (Intersection of convex sets - conceptual)
    theorems:\\
      - Sylvester-Gallai Theorem (Given n points, not all collinear, there is a line 
      containing exactly two of them)\\
      - Pick's Theorem (Area of a simple polygon whose vertices are integer lattice 
    points)\\
     \# Erdos-Anning Theorem
     \# De Bruijn–Erdős theorem (projective plane version)
 
Probability:\\
  Foundations and Basic Computations:\\
    tools\_concepts:\\
      - Sample Space, Event Space, Probability Measure (Axioms of Probability)\\
      - Equally Likely Outcomes (Classical Definition: Favorable / Total)\\
      - Complementary Events ($P(A') = 1 - P(A)$)\\
      - Union and Intersection of Events ($P(A \cup B)$, $P(A \cap B)$)\\
      - Conditional Probability ($P(A|B) = P(A \cap B) / P(B)$)\\
      - Multiplication Rule for Conditional Probability ($P(A \cap B) = P(A|B)P(B)$)\\
      - Independent Events ($P(A \cap B) = P(A)P(B)$)\\
      - Law of Total Probability (Partitioning sample space, 
      $P(A) = \sum P(A|B_i)P(B_i)$)\\
      - Bayes' Theorem\\
      - Tree Diagrams (For visualizing multi-stage experiments)
    theorems:\\
      - Boole's Inequality / Union Bound ($P(\cup A_i) \le \sum P(A_i)$)\\
      - Bonferroni Inequalities (Generalizations of PIE for probability)\\
 
  Discrete Random Variables and Expectation:\\
    tools\_concepts:\\
      - Random Variable (Definition, discrete case)\\
      - Probability Mass Function (PMF)\\
      - Expected Value (Expectation) of a Discrete Random Variable ($E[X] = 
      \sum x P(X=x)$)\\
      - Linearity of Expectation ($E[X+Y] = E[X]+E[Y]$, $E[cX] = cE[X]$ - powerful
        even if X, Y not independent)\\
      - Indicator Variables (Bernoulli random variables, $I_A=1$ if A occurs, 
        0 otherwise; $E[I_A]=P(A)$)\\
      - Variance and Standard Deviation (Basic definitions, properties like 
        $Var(X) = E[X^2] - (E[X])^2$)\\
      - Covariance (For $Var(X+Y)$, conceptual)\\
      - Common Discrete Distributions (Bernoulli, Binomial, Geometric, 
        Hypergeometric - recognizing their structure in problems)
    theorems:\\
     \# Linearity of Expectation itself is a theorem-level concept in its 
      application.\\
     \# Formulas for E[X] and Var(X) for common distributions (e.g., E[Bin(n,p)]=np)\\
 
  Combinatorial Probability (Heavy Overlap with Combinatorics Branch):
    tools\_concepts:\\
      - Counting Techniques (Permutations, combinations, stars and bars, PIE) 
      applied to find favorable/total outcomes.\\
      - Problems involving Selections, Arrangements, Distributions with 
      probabilistic questions.\\
      - Derangements (Probability that no item is in its original position).\\
      - Matching Problems (e.g., Hat-check problem).\\
      - Probabilistic arguments for existence (If $P(\text{Property holds}) > 0$, 
      then an object with that property exists).\\
    theorems:\\
     \# Many "theorems" are combinatorial identities used in a probabilistic context.\\
     \# e.g., using Vandermonde's Identity to sum probabilities in hypergeometric 
      settings.\\
 
  Geometric Probability:\\
    tools\_concepts:\\
      - Calculating probabilities as ratios of lengths, areas, or volumes.\\
      - Problems involving random points in geometric figures (intervals, squares,
      circles, cubes).\\
      - Buffon's Needle Problem (Classic example - conceptual understanding).\\
      - Coordinate Geometry methods to define regions and calculate areas/volumes.\\
      - Symmetry arguments to simplify calculations.\\
    theorems:\\
     \# Results are usually derived from geometric formulas rather than specific 
      "probability theorems" for geometry.\\
     \# Bertrand's Paradox (Illustrates importance of defining "random" precisely - 
      conceptual)\\
 
  Stochastic Processes\& Recurrence Relations (Introductory/Discrete):
    tools\_concepts:\\
      - Random Walks (1D, 2D - simple cases, probability of reaching a state, 
      expected number of steps).\\
      - Markov Chains (Finite state space, transition probabilities, transition 
      matrix - basic concepts).\\
      - First Step Analysis (Setting up recurrence relations for probabilities or 
      expected values in processes).\\
      - Gambler's Ruin Problem (Classic example).\\
      - Waiting Time Problems (e.g., expected number of trials until first success -
      Geometric distribution).\\
      - Absorption Probabilities and Expected Time to Absorption 
      (for simple Markov chains).\\
    theorems:\\
     \# Solutions to standard recurrence relations for probabilities/expectations.\\
     \# Existence of steady-state distributions for certain types of Markov chains
      (conceptual).\\
 
  Advanced Probabilistic Techniques (Rare, but can inspire hard problems):\\
    tools\_concepts:\\
      - Probabilistic Method (Using probability to prove deterministic combinatorial 
      results - beyond basic existence).\\
      - Alteration Method (Refining a random construction).\\
      - Second Moment Method (Using variance to show concentration around the 
      mean - very advanced for Olympiads).\\
      - Martingales (Sequence of random variables where conditional expectation of 
      next given current is current - highly advanced, but simplest forms/ideas 
      might inspire).\\
    theorems:\\
      - Markov's Inequality ($P(X \ge a) \le E[X]/a$)\\
      - Chebyshev's Inequality ($P(|X-\mu| \ge k\sigma) \le 1/k^2$)\\
      - Chernoff Bounds / Hoeffding's Inequality 
      (Exponential tail bounds - conceptual awareness, very rare).\\
      - Lovasz Local Lemma (Tool for showing existence when events are "mostly" 
      independent - very advanced).\\
\end{tcolorbox}



\end{document}